\documentclass{article}

\usepackage[preprint]{neurips_2022}

\usepackage[dvipsnames]{xcolor}
\definecolor{linkColor}{rgb}{0.18,0.39,0.62}
\PassOptionsToPackage{numbers, compress}{natbib}
\usepackage[utf8]{inputenc} 
\usepackage[T1]{fontenc}    
\usepackage[colorlinks=true,linkcolor=linkColor,citecolor=linkColor,filecolor=linkColor,urlcolor=linkColor]{hyperref}        
\usepackage{url}            
\usepackage{booktabs}       
\usepackage{amsfonts}       
\usepackage{nicefrac}       
\usepackage{microtype}      
\usepackage{xcolor}         
\usepackage{graphicx}
\usepackage{wrapfig}
\usepackage{amsmath}
\usepackage{arydshln}

\usepackage{newfloat}
\usepackage{hyperref}
\usepackage{makecell}

\usepackage{graphicx}
\usepackage{colortbl}
\usepackage{arydshln}
\usepackage{tikz}
\usepackage{pgfplots}
\usepackage{latexsym}
\usepackage{pifont}
\newcommand{\cmark}{\ding{51}}
\newcommand{\xmark}{\ding{55}}

\usepackage{enumitem}

\usepackage{multicol}  
\usepackage{multirow}

\newcommand{\sqlcorrect}[1]{{\color{olive}{\cmark}}}
\newcommand{\sqlwrong}[1]{{\color{red}{\xmark}}}

\title{Large Language Models are Versatile Decomposers: Decompose Evidence and Questions for Table-based Reasoning}

\author{
Yunhu Ye$^{1, 2, \includegraphics[scale=0.02]{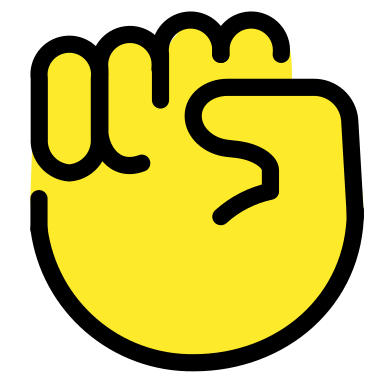}}$\footnotemark[3], Binyuan Hui$^{3, \includegraphics[scale=0.02]{first.png}}$, Min Yang$^{2}$\footnotemark[2], Binhua Li$^{3}$, Fei Huang$^{3}$, Yongbin Li$^{3}$\footnotemark[2] \\
$^1$ University of Science and Technology of China \\
$^2$ Shenzhen Institute of Advanced Technology, Chinese Academy of Sciences \\
$^3$ DAMO Academy, Alibaba Group \\
\texttt{yeyunhu@mail.ustc.edu.cn}, \texttt{min.yang@siat.ac.cn} \\
\texttt{\{binyuan.hby, shuide.lyb\}@alibaba-inc.com}\\
\url{https://github.com/AlibabaResearch/DAMO-ConvAI}
}

\begin{document}

\maketitle

\renewcommand{\thefootnote}{\fnsymbol{footnote}}
\footnotetext{\includegraphics[scale=0.02]{first.png} Equal contribution.}
\footnotetext[3]{Work done during an intern at Alibaba DAMO Academy.}
\footnotetext[2]{Corresponding authors.}

\begin{abstract}
Table-based reasoning has shown remarkable progress in a wide range of table-based tasks.
It is a challenging task, which requires reasoning over both free-form natural language (NL) questions and (semi-)structured tabular data. However, previous table-based reasoning solutions usually suffer from significant performance degradation on ``huge'' evidence (tables). In addition, most existing methods struggle to reason over complex questions since the essential information is scattered in different places. To alleviate the above challenges, we exploit large language models (LLMs) as decomposers for effective table-based reasoning, which (i) decompose huge evidence (a huge table) into sub-evidence (a small table) to mitigate the interference of useless information for table reasoning, and (ii) decompose a complex question into simpler sub-questions for text reasoning. First, we use a powerful LLM to decompose the evidence involved in the current question into the sub-evidence that retains the relevant information and excludes the remaining irrelevant information from the ``huge'' evidence. 
Second, we propose a novel ``parsing-execution-filling'' strategy to decompose a complex question into simper step-by-step sub-questions by generating intermediate SQL queries as a bridge to produce numerical and logical sub-questions with a powerful LLM.
Finally, we leverage the decomposed sub-evidence and sub-questions to get the final answer with a few in-context prompting examples. 
Extensive experiments on three benchmark datasets (TabFact, WikiTableQuestion, and FetaQA) demonstrate that our method achieves significantly better results than competitive baselines for table-based reasoning. 
Notably, \textbf{our method outperforms human performance for the first time on the TabFact dataset}. In addition to impressive overall performance, our method also has the advantage of interpretability, where the returned results are to some extent tractable with the generated sub-evidence and sub-questions.
\end{abstract}

\section{Introduction}

\begin{wrapfigure}{l}{0.5\textwidth}
    \centering
    \setlength{\abovecaptionskip}{5pt}
    \includegraphics[width=0.5\textwidth]{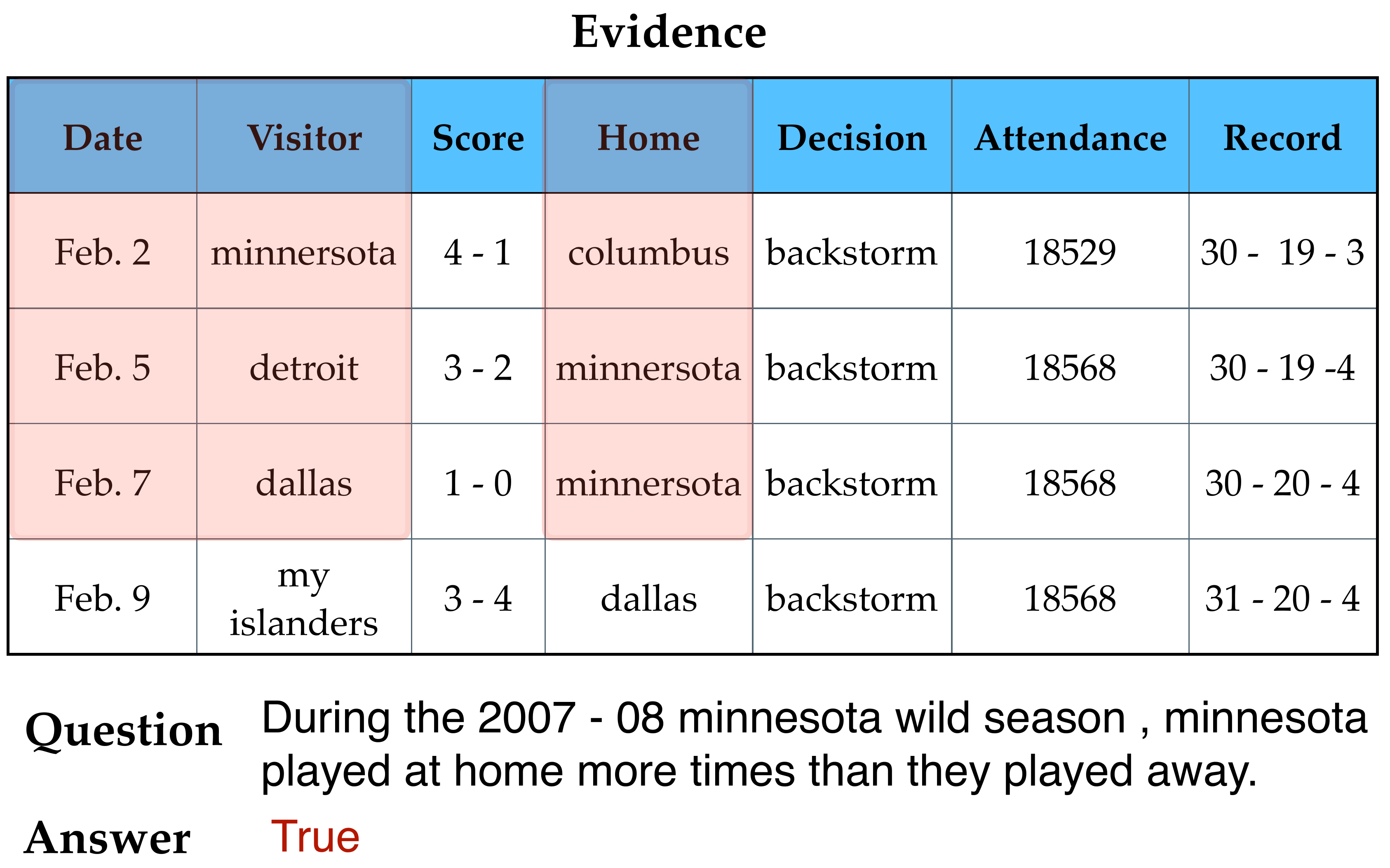}
    \caption{An example of table-based reasoning.}
    \label{intro}
\end{wrapfigure}

Tabular data can be an important complement to textual data, which is informative and ubiquitous in our daily lives. Reasoning about tabular and textual information is a fundamental problem in natural language understanding (NLU) and information retrieval (IR) \citep{wang2021retrieving}. 
It can benefit various downstream applications such as table-based fact verification (FV) \citep{tabfact,feverous,infotabs} and table-based question answering (QA) \citep{wtq,wikisql,fetaqa,cho2019sigirtable}. As shown in Figure~\ref{intro}, table-based reasoning is challenging since it involves sophisticated textual, numerical, and logical reasoning across both unstructured text and (semi-)structured tables. 
To address the above challenge, table-based reasoning is conventionally accomplished by synthesizing executable languages (e.g., SQL and SPARQL) to interact with tables \citep{sqlguide,sparql,Hui2021DynamicHR,Hui2022S2SQLIS}. However, these methods ignore the semantics of text chunks inside the table, struggling to effectively model web tables with unnormalized free-form text in table cells.

Recently, table-based pre-trained models\citep{tapas,tapex,omnitab,pasta,Cai2022STARSG} have proved to be powerful in enhancing reasoning skills on both textual and tabular data, which benefit from the rich knowledge learned from large-scale crawled or synthesized tabular and textual data. For example, TaPas~\citep{tapas} enhanced the understanding of tubular data by recovering masked cell information in  tables. 
Nevertheless, these models generally need to be fine-tuned on a significant amount of task-specific downstream datasets, struggling to achieve excellent performances when dealing with new datasets with unseen reasoning types. In addition, fine-tuning a pre-trained model on specific tasks generally destroys its in-context ability \citep{wang2022preserving}. Instead of fine-tuning the pre-trained models, in-context learning in large language models (LLMs) \citep{gpt3,cot,zerocot,min2022rethinking} has recently gained noticeable attention to exploring the reasoning ability of LLMs, where several input–output exemplars are provided for prompting. For example, the \textit{chain of thought} prompting \citep{cot} has been discovered to empower LLMs to perform complex reasoning by generating intermediate reasoning steps.

The prior studies \citep{cot,zerocot} revealed that LLMs could achieve impressive performance on many text reasoning tasks without task-specific model designs and training data. However, the capability of LLMs on table-based reasoning tasks is still under-explored.  
There are several technical challenges in leveraging LLMs for table-based reasoning with huge tables and complex questions. \textbf{First}, since a table could potentially contain a large number of rows and columns, directly encoding all table content via pre-trained models could be computationally intractable and interfere with huge irrelevant information. As pointed out in \citep{llmtab}, the LLMs are unable to generalize to ``huge'' tables with 30+ rows due to the token limitation. Although previous studies have leveraged some approaches \citep{yin2020tabert,tabfact} such as text matching to retrieve sub-evidence, these approaches usually require a large amount of domain-specific data for training, as the sub-evidence retrieval process requires deep understanding and reasoning of questions and tables. 
\textbf{Second}, decomposing a complex question into simpler sub-questions can effectively facilitate multi-step reasoning of a large model \citep{huang2022language,dua2022successive,pot}. However, directly decomposing a complex question by leveraging chain-of-thought prompting \citep{cot} could easily fall into a hallucination dilemma \citep{ji2022survey}, where the models may generate misleading sub-questions containing information that is inconsistent with the evidence. 
Chen \textit{et al.} \citep{llmtab} showed that LLMs could sometimes make simple mistakes when performing symbolic operations.
This will affect the process of subsequent reasoning, thus we need a more reliable table-based method for complex question decomposition

To mitigate the aforementioned challenges, in this paper, we explore in-context learning in LLMs to \underline{D}ecompose evidence \underline{A}nd questions for effective \underline{T}able-bas\underline{E}d \underline{R}easoning (\textbf{\textsc{Dater}}). 
First, we exploit a powerful LLM to decompose the (semi-)structured evidence (a huge table) involved in the current question into relevant sub-evidence (a small table). We implement sub-evidence extraction by predicting the indexes of the rows and columns with the help of a powerful LLM and a few prompts.
In this way, we can retain the relevant evidence and exclude the remaining irrelevant evidence from interfering with the decision. In addition, our method has the advantage of interpretability, where the returned table-based results are tractable.
Second, we propose a ``\textit{parsing-execution-filling}'' strategy which explores the programming language SQL to decompose the complex unstructured natural language (NL) question into the logical and numerical computation. Concretely, we generate an abstract logic sub-question by masking the span of numerical values and then converting the abstract logic into SQL query language that is executed on the evidence for obtaining a reliable sub-question. Finally, we leverage the decomposed sub-evidence and sub-questions to get the final answer with the help of a few in-context prompting examples.  

The main contributions of this paper are listed as follows:
\begin{itemize}
    \item We reduce ``huge'' evidence (a huge table) into ``small'' sub-evidence (a small table) by predicting the related indexes of rows and columns of the evidence with the help of a powerful LLM and a few prompting examples. Our evidence decomposition method makes the reasoners focus on the essential sub-evidence related to the given question.
    \item We propose a novel ``parsing-execution-filling'' strategy to decompose a complex question into simper step-by-step sub-questions by generating an intermediate SQL as a bridge to produce numerical and logical sub-questions with the help of a powerful LLM and a few prompting examples. Our question decomposition method has proven to be effective in table-based reasoning without requiring a large amount of annotated training data. 
    \item We conduct extensive experiments on three benchmark datasets belonging to table-based fact verification and table-based question answering tasks. Experimental results demonstrate that our Dater method achieves significantly better results than competitive baselines for table-based reasoning. Particularly, Dater outperforms human performance for the first time on the TabFact dataset. 
    \item In addition to impressive overall performance, our Dater also has the advantage of interpretability, where the returned results are to some extent tractable with the generated sub-evidence and sub-questions.
\end{itemize}

\section{Related Work}
\subsection{Table-based Reasoning}
Table-based reasoning requires reasoning over both free-form natural language (NL) questions and (semi-)structured tables. Traditional methods produce executable languages (e.g., SQL and SPARQL) to access the tabular data \citep{sqlguide,sparql}. However, these methods cannot capture the semantics of text chunks inside a table and fail to model web tables with free-form text in table cells. 
Recently, several table-based reasoning benchmarks, such as TabFact \citep{tabfact}, WikiTableQuestion \citep{wtq}, and FetaQA \citep{fetaqa},  have been proposed to help learn different types of table-based tasks.  The availability of large-scale training data has significantly enhanced the performance of table-based reasoning with the help of deep learning techniques \citep{neeraja2021incorporating,feverous}.  

In parallel, table pre-training has been proposed to encode both tables and texts, which further improves the performance of table-based reasoning. 
Inspired by the success of masked language modeling (MLM), TaPas~\citep{tapas} enhanced the understanding of tubular data by recovering masked cell information in the table. TAPEX~\citep{tapex} utilized the BART model to imitate the SQL executor in the pre-training stage so that TAPEX can obtain better table reasoning capability. ReasTAP~\citep{reastap} designed pre-training tasks based on the reasoning skills of table-based tasks, injecting reasoning skills via pre-training. 
\textsc{TaBERT} \citep{yin2020tabert} proposed content snapshots to encode a subset of table content that was most relevant to the input utterance. 
PASTA~\citep{pasta} introduced a table-operations aware
fact verification approach, which pre-trained LMs to be aware of common table-based operations with sentence-table cloze questions synthesized from WikiTables.
Subsequently, \citep{llmtab,binder} explored the ability of LLMs for table-based tasks.

\subsection{Large Language Models on Reasoning}
Large language models (LLMs) have been shown to confer a range of reasoning abilities, such as arithmetic \citep{lewkowycz2022solving}, commonsense \citep{liu2022rainier} and symbolic reasoning \citep{zhou2022least}, as the model parameters are scaled up \citep{gpt3}.
Notable, chain-of-thought (CoT) \citep{cot} leverages a series of intermediate reasoning steps, achieving better reasoning performance on complex tasks. 
Based on CoT, a number of advanced improvements have been proposed, including ensemble process \citep{cotsc}, iterative optimization \citep{zelikman2022star}, and example selection \citep{creswell2022selection}.
Notably, ZeroCoT \citep{zerocot} improved reasoning performance by simply adding ``Let's think step by step'' before each answer.
Fu \textit{et al.} \citep{fu2022complexity} proposed complexity-based prompting can generate more reasoning steps for the chain, and achieve significantly better performance.
Zhang \textit{et al.} \citep{zhang2022automatic} selected examples of in-context automatically by clustering, without the need for manual writing.
Despite the remarkable performance of LLMs in textual reasoning, their reasoning capabilities on tabular tasks are still limited.
The two most relevant works to this paper are  \citep{binder} and \citep{llmtab}, but none of them focus on the ability of a powerful LLM to decompose the evidence (tables), and the reliability of the reasoning steps.

\begin{figure*}
    \centering
    \setlength{\abovecaptionskip}{5pt} 
    \includegraphics[width=0.9\textwidth]{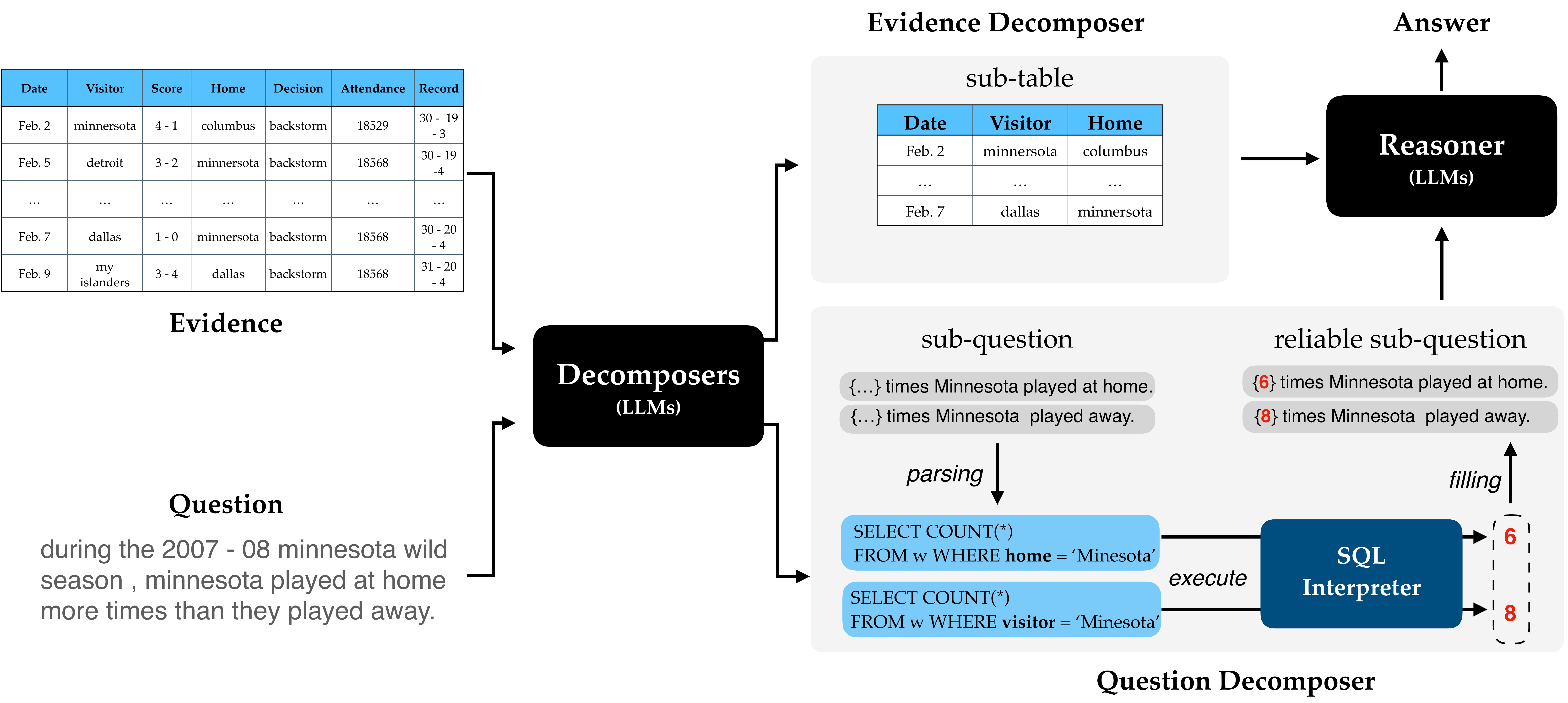}
    \caption{The overview of our Dater framework for table-based reasoning. We first use a powerful LLM (Codex) to probe sub-evidence and sub-questions by performing in-context learning. To obtain a reliable sub-question, we propose a novel ``parsing-execution-filling'' strategy to alleviate hallucination issues. Ultimately, the reasoner browses through the sub-evidence and sub-questions to get the final answer.
    }
    \label{pipeline}
\end{figure*}

\subsection{Question Decomposition}
Question decomposition is essential to understand complex questions. Early work \citep{kalyanpur2012fact} utilized a suite of decomposition rules for question decomposition based on lexicon-syntactic features. The drawback of rule-based methods is that they need experts to manually design rules, making it difficult to extend the rules to new tasks or domains. In recent years, neural models \citep{talmor2018web,zhang2019complex} have been proposed to perform question decomposition in an end-to-end manner.
Zhang \textit{et al.} \citep{zhang2019complex} proposed a hierarchical semantic parsing method based on a sequence-to-sequence model, which combined a question decomposer and an information extractor. 
However, these supervised methods rely on a large amount of annotated training data which is labor-intensive to obtain. In parallel, unsupervised decomposition was proposed to produce sub-questions without strong supervision. For example, Perez \textit{et al.} \citep{perez2020unsupervised} automatically produced a noisy ``pseudo-decomposition'' for
each complex question, which trained a
decomposition model on the crawled data with unsupervised sequence-to-sequence learning. 
Different from previous decomposition methods, we explore LLMs as versatile decomposers, which decompose both huge evidence and complex questions for table-based reasoning with the help of a powerful LLM and a few prompting examples.

\section{Problem Formulation and Notations}
Each instance in table-based reasoning consists of a table $T$, a natural language question $Q$, and an answer $A$. In particular, each table $T=\{v_{i,j}|i<= {Row}_T,j<={Col}_T\}$ contains ${Row}_T$ rows and ${Col}_T$ columns, with $v_{i,j}$ representing the content in the $(i, j)$-th cell. A question $Q = <q_1,{\cdots} ,q_n >$ consists of $n$ tokens. In this paper, we focus on two table-based reasoning tasks, including table-based fact verification (FV) and table-based question answering (QA). For table-based FV, the final answer $A \in \{0, 1\}$ is a boolean value that determines the truth or falsity of the input statement. For table-based QA, the answer is a natural language sequence $A = <a_1,{\cdots}, a_n> $ that answers the question described by the input statement.

\section{Method}
\subsection{In-context Learning}
Given a few examples with instructions about detailed specific tasks, Large Language Models (LLMs) can perform in-context learning without training, which learn from analogies.
The powerful ability has been widely verified on nature language tasks, including text classification, semantic parsing, mathematical reasoning, etc.
Inspired by the advances of LLMs, we aim to explore whether LLMs could tackle reasoning tasks on structured evidence (i.e., tabular data).
Formally, the final answer $A_{\text{test}}$ can be obtained by predicting $p(A_{\text{test}} \mid T_{\text{test}}, Q_{\text{test}}, C)$ with evidence table $T_{\text{test}}$ and question $Q_{\text{test}}$. 
Here, $C=\left\{C_1, \ldots, C_{|C|}\right\}$ is a small set of in-context prompts from manual writing, where each example $C_i = (T_{\text{prompt}}^i, Q_{\text{prompt}}^i, A_{\text{prompt}}^i)$. 
We provide the detailed in-context prompt as prompt 4.1.

\begin{figure}[htbp]
\centering
\begin{minipage}[t]{0.48\textwidth}
\centering
\includegraphics[width=6cm]{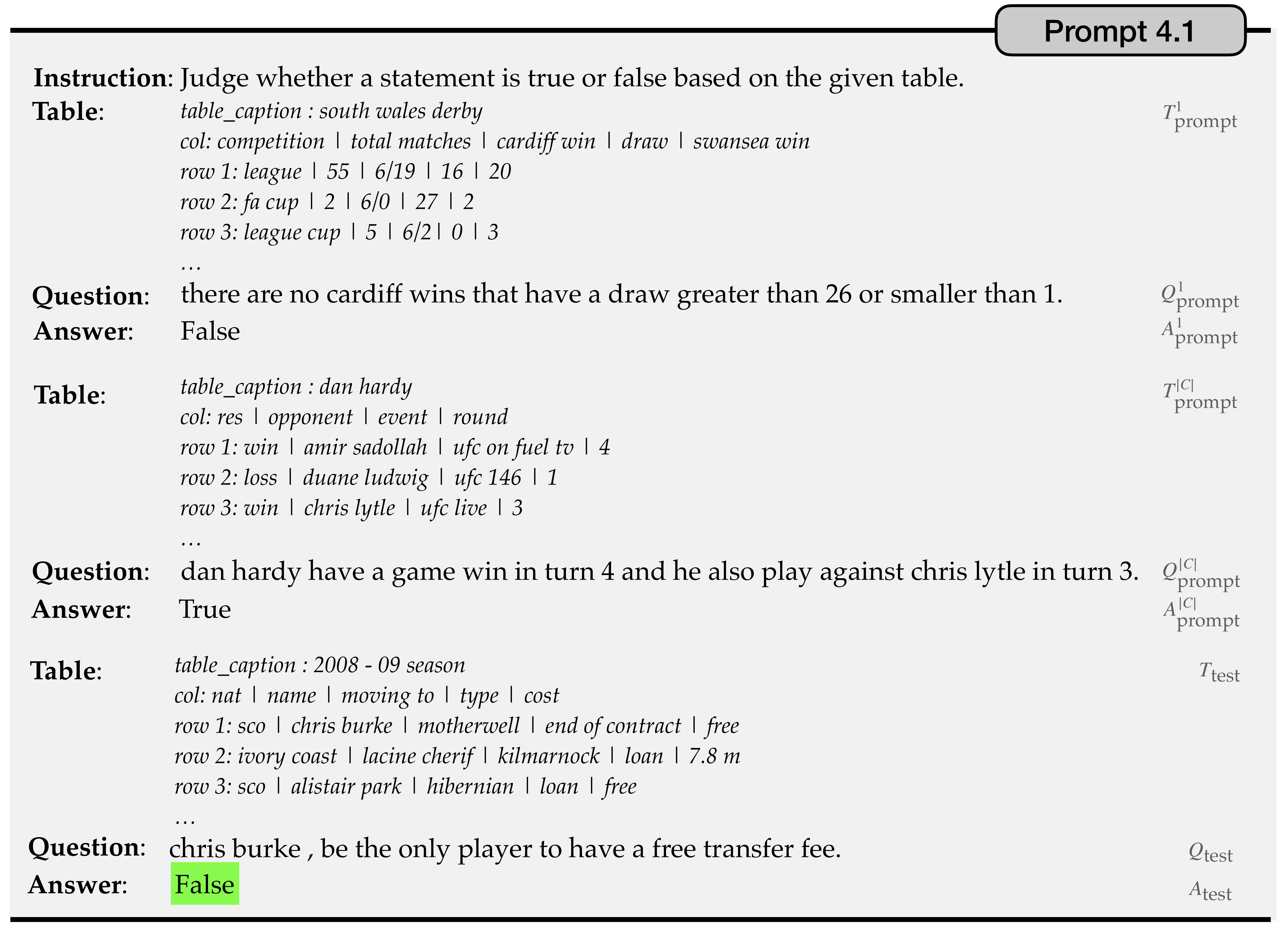}
\end{minipage}
\begin{minipage}[t]{0.48\textwidth}
\centering
\includegraphics[width=6cm]{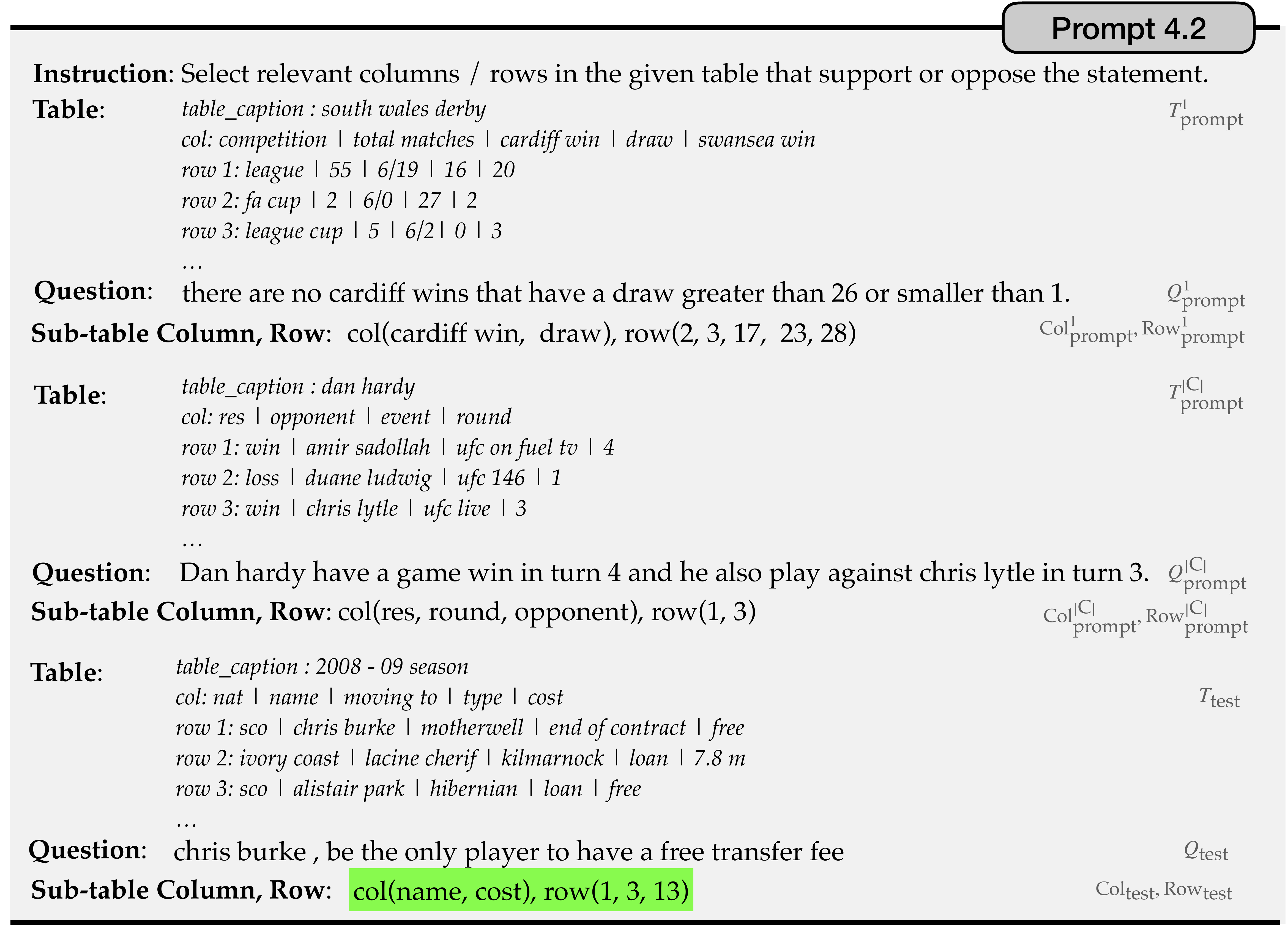}
\end{minipage}
\end{figure}

Based on empirical observations, performing in-context learning directly by the above prompt 4.1 cannot achieve optimal performance. Table-based reasoning is a sophisticated and complex task that requires human-like thinking and a fine-grained understanding process. Recently, chain-of-thought \citep{cot} proposed a question decomposition method, which induces LLMs to perform step-by-step thinking, resulting in better reasoning. 
However, they did not break down the evidence, resulting in unsatisfactory performance. Specifically, dealing with the ponderous tables directly is usually inefficient and easily interfered with by huge irrelevant information. 
To this end, we introduce evidence and question decomposition to help large models perform fine-grained reasoning over both evidence (tables) and questions.

\subsection{Evidence Decomposer}
Humans process table-based reasoning tasks by observing sub-evidence related to the question at hand to complete their judgment. In this paper, we hope to use evidence decomposition to imitate this process.
Although previous studies have leveraged some approaches \citep{yin2020tabert,tabfact} such as text matching to retrieve sub-evidence, empirical results suggest that these approaches are often imperfect and require a large amount of domain-specific data for training, as the sub-evidence retrieval process relies on strong commonsense as well as domain knowledge and requires a joint understanding and reasoning of questions and tables.
To this end, we use a powerful LLM to break down the evidence (tables) involved in the current question, retaining the relevant evidence and excluding the remaining irrelevant evidence from interfering with the decision. 
Concretely, we implement sub-evidence extraction by predicting the indexes of the rows and columns with the help of a powerful LLM and a few prompts.
Formally, in the in-context learning stage, the row index $\text{Row}_{\text{test}} = \{\text{Row}_1, \text{Row}_2, ... \text{Row}_{|\text{Row}|}\}$ and column index $\text{Col}_{\text{test}} = \{\text{Col}_1, \text{Col}_2, ... \text{Col}_{|\text{Col}|}\}$ of sub-evidence table $\hat{T}_{\text{test}}$ can be obtained by predicting $p(\text{Row}_{\text{test}}, \text{Col}_{\text{test}} \mid T_{\text{test}}, Q_{\text{test}}, C^\text{ED})$ with complete evidence $T_{\text{test}}$ and corresponding question $Q_{\text{test}}$. $C^{\text{ED}}=\left\{C^\text{ED}_1, \ldots, C^\text{ED}_{|\text{C}|}\right\}$ is a small set of in-context examples, where each $C^\text{ED}_i$ is the example instance $(\text{Row}_\text{prompt}^i, \text{Col}_\text{prompt}^i, T_{\text{prompt}}^i, Q_{\text{prompt}}^i)$. Some detailed prompts are illustrated as prompt 4.2.

\subsection{Questions Decomposer}
Decomposing complex questions into step-by-step sub-questions can effectively facilitate the reasoning process of a large model, which has been proven to be effective in numerical and commonsense reasoning \citep{huang2022language,dua2022successive,pot}. However, we observed that 
straightforwardly decomposing a complex question by leveraging a chain-of-thought process could easily fall into a hallucination dilemma, i.e., the LLM might not faithfully generate content consistent with the given evidence (tables), especially when numerical values were involved. This will affect the process of subsequent reasoning, thus we need a reliable method for the sub-question generation.

\subsubsection{Reliable Sub-questions}
To effectively decompose a complex question into step-by-step sub-questions, we propose a ``\textbf{parsing-execution-filling}'' strategy to extend the vanilla chain-of-thought method by exploring the programming language SQL to divide logical steps and numerical computation. 
Specifically, we first generate an abstract logic sub-question, where we mask the span of numerical values by using cloze style and then convert the abstract logic into SQL queries, similar to text-to-SQL parsing \citep{Qin2022ASO,Wang2022ProtonPS}. 
Afterward, we execute the SQL language on the evidence to get the final result for back-filling, yielding a reliable sub-question.
For example, as shown in the bottom of Figure \ref{pipeline}, given a question 
``\textit{during the 2007 - 08 minnesota wild season, minnesota played at home more times than they played away.}''
, we first mask the spans involving numerical values in sub-questions in prompting examples, and the remaining part can be regarded as the logical question. Here, the logical sub-questions are 
``\textit{q1: \{...\} times minnesota played at home.}'' and ``\textit{q2: \{...\} times minnesota played away}.'' Formally, a sub-question $\hat{Q}_{\text{test}}$ can be obtained by predicting $p(\hat{Q}_\text{test} \mid T_{\text{test}}, Q_{\text{test}}, C^{\text{QD}})$ with the complete evidence $T_{\text{test}}$ and complete question $Q_{\text{test}}$. 
$C^{\text{QD}}=\left\{C^{\text{QD}}_1, \ldots, C^{\text{QD}}_{|\text{C}|}\right\}$ is a small set of in-context prompting examples, where each $C^{\text{QD}}_i$ is an example instance $(T_{\text{prompt}}^i, Q_{\text{prompt}}^i, \hat{Q}_{\text{prompt}}^i)$.
The detailed prompt is as prompt 4.3.
\begin{figure}[htbp]
\centering
\begin{minipage}[t]{0.48\textwidth}
\centering
\includegraphics[width=6cm]{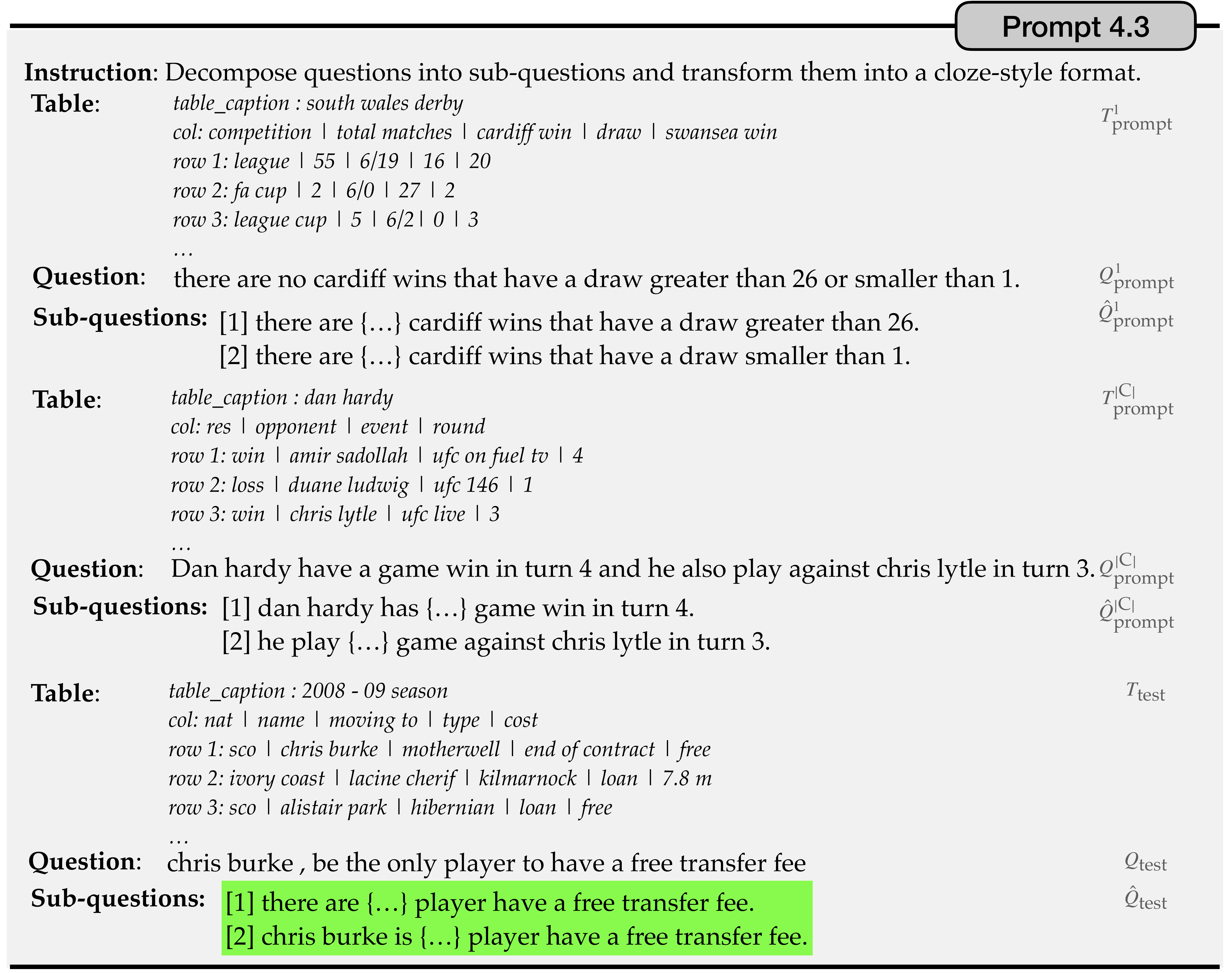}
\end{minipage}
\begin{minipage}[t]{0.48\textwidth}
\centering
\includegraphics[width=6cm]{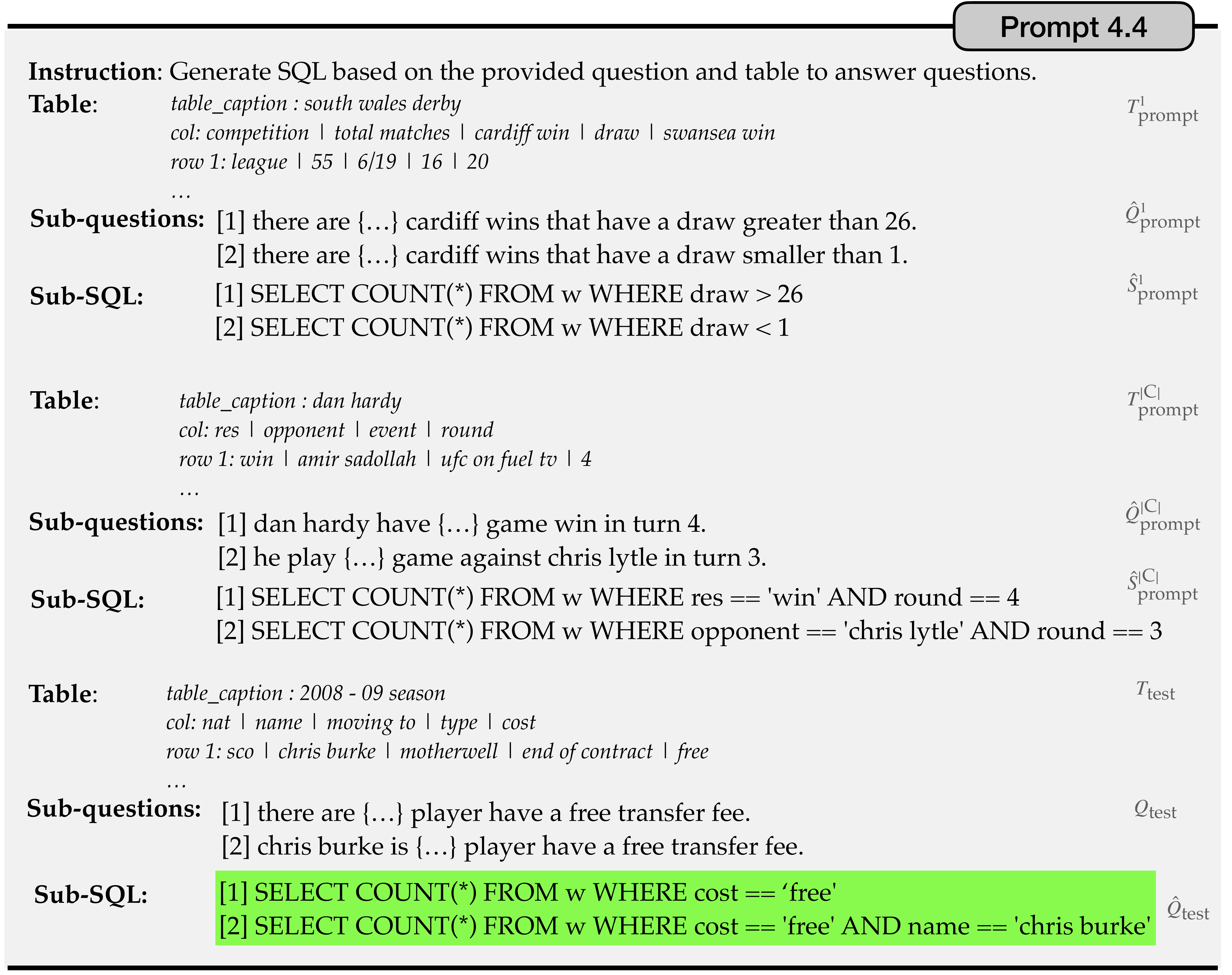}
\end{minipage}
\end{figure}

The logical statement is then parsed into the SQL queries ``\texttt{SELECT COUNT(*) FROM w WHERE home = 'minnesota'}'' and 
``\texttt{SELECT COUNT(*) FROM w WHERE visitor = 'minnesota'}'',
which are further executed on the evidence, yielding reliable spans ``\textit{6}'' and ``\textit{8}'' to be backfilled into the placeholder $\{...\}$ of sub-questions. In this way, we can obtain the reliable sub-questions 
``\textit{q1: \{6\} times minnesota played at home.}''
and ``\textit{q2: \{8\} times minnesota played away.}''. 

Formally, the SQL query $\hat{S}_{\text{test}}$ can be obtained by predicting $p(\hat{S}_\text{test} \mid T_{\text{test}}, \hat{Q}_{\text{test}}, C^{\text{PR}})$ with the help of complete evidence $T_{\text{test}}$ and sub-question $Q_{\text{test}}$. Here, 
$C^{\text{PR}}=\left\{C^{\text{PR}}_1, \ldots, C^{\text{PR}}_{|\text{C}|}\right\}$ is a small set of in-context examples, where each $C^{\text{PR}}_i$ is an example instance $(T_{\text{prompt}}^i, \hat{Q}_{\text{prompt}}^i, \hat{S}_{\text{prompt}}^i)$.
The detailed prompt is as prompt 4.4.

\subsection{Jointly Reasoning}
After performing the above evidence and question decomposition, the reasoner leverages both sub-evidence $\hat{T}_{\text{test}}$ and reliable sub-questions $\hat{Q}_{\text{test}}$ to get a final answer $A_{\text{test}}$ by predicting $p(A_{\text{test}} \mid \hat{T}_{\text{test}}, \hat{Q}_{\text{test}}, C^{\text{JR}})$, where $C^{\text{JR}}=\left\{C^{\text{JR}}_1, \ldots, C^{\text{JR}}_{|\text{C}|}\right\}$ is a small set of new in-context prompting examples. Here, each prompting example is denoted as $C^{\text{JR}}_i = (\hat{T}_{\text{prompt}}^i, \hat{Q}_{\text{prompt}}^i, A_{\text{prompt}}^i)$.
The detailed prompt as prompt 4.5.
\begin{figure}
    \centering
    \includegraphics[width=0.5\textwidth]{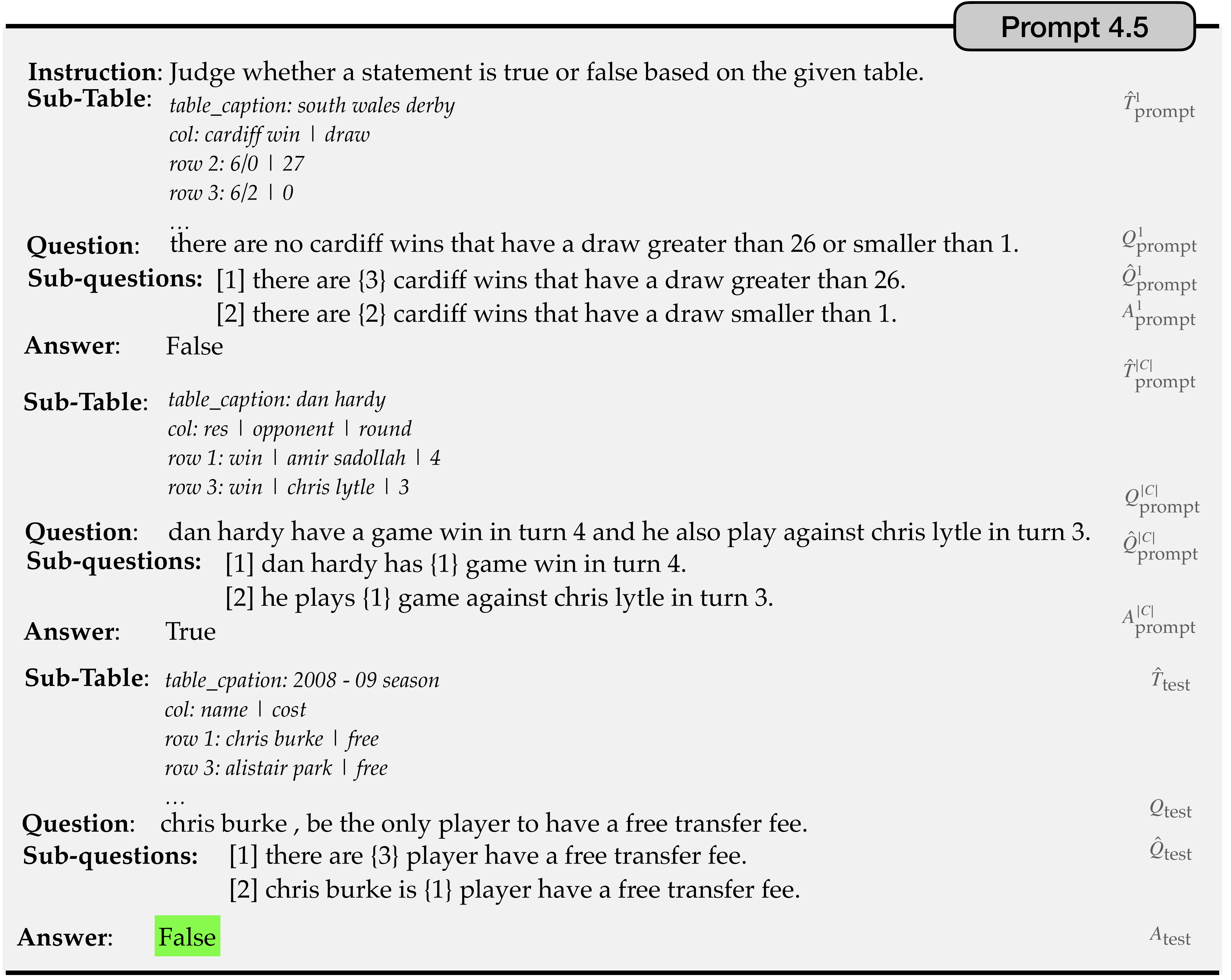}
    \label{p45}
\end{figure}

\section{Experimental Setup}

\subsection{Datasets}
We evaluate our proposed method Dater on three table-based reasoning datasets, including TabFact \citep{tabfact}, WikiTableQuestion \citep{wtq}, and FetaQA \citep{fetaqa}. 
Given the request and cost constraints of the LLMs, we only evaluate our Dater on the test sets of these corpora, without fine-tuning it on the training sets. 
It is noteworthy that LLMs were trained primarily on the web-crawled and code data. 
Since the pre-training data of LLMs does not contain tabular data, there is no risk of test data leakage. 
The details of these three datasets are provided as follows.

\begin{itemize}
    \item \textbf{TabFact} is a table-based fact verification benchmark, which contains statements written by crowd workers based on Wikipedia tables. For example, a \textit{statement: ``industrial and commercial panel has four more members than the cultural and educational panel.''} is need to be judged whether it is \textit{``True''} or \textit{``False''} according to the given table.
    We report the accuracy on the test-small set containing 2,024 statements with 298 tables.
    \item \textbf{WikiTableQuestion} contains complex questions annotated by crowd workers based on Wikipedia tables. The crowd workers are asked to write questions that involve several complex operations such as comparison, aggregation and arithmetic operations, which require compositional reasoning over a series of entries in the given table. 
    We use the standard validation set and test set with 2,381 and 4,344 samples, respectively.
    \item \textbf{FetaQA} contains free-form table questions that require deep reasoning and understanding. Most questions in FetaQA are based on information from discontinuous chunks in the table. 
    We evaluate Dater on the test set containing 2,003 samples. 
\end{itemize}

\subsection{Evaluation Metrics}
TabFact is used to evaluate table-based fact verification which aims to check if a statement is true based on tables. We adopt binary classification accuracy as the evaluation metric for the TabFact dataset. For WikiTableQuestion, we use denotation accuracy as our evaluation metric, which verifies whether the predicted answers equal the gold ones. Different from TabFact and WikiTableQuestion datasets that produce short phrase answers, the goal of FetaQA is to generate a complete long-form answer. Therefore, for FetaQA, we utilize BLEU \citep{bleu}, ROUGE-1, ROUGE-2, ROUGE-L \citep{rouge} as evaluation metrics.

\subsection{Implementation Details}
In the experiments, we employ GPT-3 Codex (code-davinci-002) as our large language model.  
For the final reasoning in-context learning step, we annotate 4, 2, and 6 prompting examples for TabFact, WikiTableQuestion, and FetaQA, respectively. 
To obtain consistent results, we use a self-consistence decoding strategy \citep{cotsc}.

\begin{table}
\centering
\begin{minipage}[t]{0.45\linewidth}
  \centering
  \begin{tabular}{ll}
    \toprule
    \textsc{\textbf{Model}} & \textsc{\textbf{Test}} \\
    \midrule
    \multicolumn{2}{c}{\textit{ $\heartsuit$ Fine-tuning based Methods}} \\
    Table-BERT & 68.1 \\
    LogicFactChecker & 74.3\\
    SAT     &   75.5 \\
    TaPas & 83.9\\
    TAPEX & 85.9 \\
    SaMoE & 86.7\\
    PASTA & 90.8 \\
    \rowcolor[RGB]{237,237,237} \quad w/ \textbf{\textsc{Dater}} & \textbf{93.0} ($\uparrow$ 2.2) \\
    \midrule
    \textit{\texttt{Human}}& {92.1}\\
    \midrule
    \multicolumn{2}{c}{\textit{ $\spadesuit$ LLM based Methods}} \\
    Binder & 85.1 \\
    Codex & 72.6 \\
    \rowcolor[RGB]{237,237,237} \quad w/ \textbf{\textsc{Dater}} & \textbf{85.6} ($\uparrow$ 13.0)  \\
    \bottomrule
  \end{tabular}
  \vspace{0.2cm}
  \caption{Experimental results on the official small test set of TabFact. Here, ``\textit{Human}'' indicates the human performance.}
  \label{tab:tabfact_small}
\end{minipage}
\hspace{0.3cm}
\begin{minipage}[t]{0.5\linewidth}
  \centering
  \begin{tabular}{lcc}
    \toprule
    \textsc{\textbf{Model}} & \textsc{\textbf{Dev}} & \textsc{\textbf{Test}} \\
    \midrule
    \multicolumn{3}{c}{\textit{$\heartsuit$ Fine-tuning based Models}} \\
    MAPO & 42.7 & 43.8 \\
    MeRL & 43.2 & 44.1 \\
    LatentAlignment & 43.7 & 44.5\\
    IterativeSearch & 43.1 & 44.7\\
    T5-3B  & - & 49.3 \\
    TaPas  & 49.9 & 50.4 \\
    TableFormer & 51.3 & 52.6 \\
    TAPEX  & 58.0 & 57.2 \\
    ReasTAP & 58.3 & 58.6\\
    TaCube  & 60.9 & 60.8 \\
    OmniTab &  61.3 & 61.2 \\
    \rowcolor[RGB]{237,237,237} \quad w/ \textbf{\textsc{Dater}} & \textbf{62.5} ($\uparrow$ 1.2)\ & \textbf{62.5} ($\uparrow$ 1.3)\\
    \midrule
        \multicolumn{3}{c}{\textit{ $\spadesuit$ LLM based Methods}} \\
    Binder & 62.6 & 61.9 \\
    Codex  & 49.3 & 47.6 \\
    \rowcolor[RGB]{237,237,237} \quad w/ \textbf{\textsc{Dater}} & \textbf{64.8}($\uparrow$ 15.5)\  & \textbf{65.9} ($\uparrow$ 18.3) \\
    \bottomrule
  \end{tabular}
  \vspace{0.2cm}
  \caption{
  Experimental results on WikiTableQuestion with the official evaluator. 
  }
  \label{tab:wikitq}
\end{minipage}
\end{table}

\subsection{Baselines}
We compare the proposed Dater with a range of strong baseline methods that can be divided into two categories: fine-tuning based methods that require training on task-specific data and LLM-based in-context learning methods that do not require fine-tuning.

\paragraph{$\heartsuit$ \textbf{Fine-tuning based Methods}} 
For the fine-tuning based methods, we compare the following methods:
\textbf{Table-BERT} \citep{tabfact} adopts rule-based method to linearize a table into a natural language (NL) sentence. The linearized table and statement are then directly encoded using a BERT model.
\textbf{LogicFactChecker} \citep{logicalfactchecker} employs a sequence-to-action semantic parser to generate a program, represented as a tree with multiple operations. A graph neural network is then utilized to encode statements, tables, and the generated programs.
\textbf{TaPas} \citep{tapas} enhances the BERT architecture by incorporating the ability to encode tables as input. It is pre-trained on a joint dataset of text segments and tables collected from Wikipedia, allowing the model to better process tabular data and improving the overall performances on various downstream tasks.
\textbf{SAT} \citep{sat} improves the model's ability to attend to the relevant information within a table by introducing a structure-aware mask matrix. In particular, it focuses on capturing low-level lexical information in the low layers and semantic information in the upper layers.
\textbf{TAPEX} \citep{tapex} is designed to make the BART model imitate the SQL executor, which allows the model to acquire better table comprehension ability. The corpus of pre-trained SQL query templates are extracted from the SQUALL dataset. 
\textbf{SaMoE} \citep{samoe} introduces mixture-of-experts (MoE) \citep{moesurvey} into the field of table-based fact verification, aiming to make different experts focus on different types of reasoning tasks. 
\textbf{ReasTAP} \citep{reastap} defines seven reasoning skills over semi-structured data, including in conjunction, temporal comparison, date difference, and so on. It infused tabular reasoning capability by pre-training the model on synthetic dataset.
\textbf{PASTA} \citep{pasta} designs six types of sentence–table cloze tasks that are pre-trained on a synthesized corpus of 1.2 million items from WikiTables. The ``\textit{fine-tuning with select-then-rank}'' strategy is used to adapt the input length to the model, which allows the model to handle long input sequences effectively. 
\textbf{TableFormer} \citep{tableformer} proposes a method that is robust to perturbations in table rows and columns, thereby improving the model's understanding of tables through the use of positional encoding.
\textbf{TaCube} \citep{tacube} is a pre-computation-based approach aiming to improve the ability of PLMs in numerical reasoning. This method pre-computes aggregation/arithmetic results for the table in advance, making them readily available for PLMs when tackling question answering.
\textbf{OmniTab} \citep{omnitab} proposes an omnivorous pre-training approach that consumes both natural and synthetic data to enhance models with abilities for both types of data.

\paragraph{$\spadesuit$ \textbf{LLM based Methods}} For the LLM based methods with in-context learning, we compare the following methods:
\textbf{Codex} \citep{codex} directly generates final answer by performing in-context learning as shown in Prompt 4.1.  
\textbf{Binder} \citep{binder} generates programming language programs and extends the capability of the programming language to solve commonsense problems.

\begin{table}
  \centering
  \begin{tabular}{lcccc}
    \toprule
    \textsc{\textbf{Model}} & \textsc{\textbf{BLEU}} &  \textsc{\textbf{ROUGE-1}} & \textsc{\textbf{ROUGE-2}} & \textsc{\textbf{ROUGE-L}} \\
    \midrule
    \multicolumn{5}{c}{\textit{$\heartsuit$ Fine-tuning based Methods}} \\
    T5-small    &   21.60   &   0.55    &   0.33    &   0.47    \\
    T5-base     &   28.14   &   0.61    &   0.39    &   0.51    \\
    T5-large    &   30.54   &   0.63    &   0.41    &   0.53    \\
    \midrule
    \multicolumn{5}{c}{\textit{ $\spadesuit$ LLM based Methods}} \\
    Codex & 27.96  &   0.62   &   0.40   &0.52      \\
    \rowcolor[RGB]{237,237,237} \quad w/ \textbf{\textsc{Dater}}   &   \textbf{30.92}    &   \textbf{0.66}   &   \textbf{0.45}    &    \textbf{0.56}    \\
    \bottomrule
  \end{tabular}
  \vspace{0.2cm}
  \caption{Evaluation on FeTaQA test set.}
  \label{tab:fetaqa}
\end{table}

\subsection{Main Results}
We conduct extensive experiments on TabFact, WikiTableQuesion and FetaQA. The experimental results on TabFact are summarized in Table~\ref{tab:tabfact_small}. From the results, we can observe that our Dater model achieves a substantial improvement over the compared methods. 
If we use LLM (Codex) as reasoner, our Dater achieves an accuracy of 85.6\%, which is 13.0\% higher than the result predicted by the Codex without evidence and questions decomposition stages. 
On the other hand, when we use the fine-tuned model as reasoner in Dater framework, i.e., injecting intermediate decomposition generated by Dater into PASTA, the accuracy increases by 2.2\% (93.0\% vs. 90.8\%) over PASTA.
It is noteworthy that \textbf{Dater outperforms human performance for the first time on the TabFact dataset}.  

Table~\ref{tab:wikitq} provides the experimental results on the WikiTableQuestion dataset. 
Dater achieves a new state-of-the-art accuracy of 65.9\% on the test set, surpassing the best baseline method (i.e., Binder) by 4.0\%.
When injecting intermediate decomposition generated by Dater into OmniTab, we can still achieve a gain of 1.3\% on the test set. 
In addition, Dater has an absolute gain of 18.3\% compared to the original Codex for table-based QA. The impressive improvement may be due to the fact that WikiTableQuestion has relatively large tables and complex questions, which can be effectively addressed by our table and question decomposition methods.
Table~\ref{tab:fetaqa} illustrates the experimental results on the FetaQA dataset.
We can observe that our Dater method achieves better results than the compared methods T5 and Codex, further validating the effectiveness of Dater.

\begin{table*}
    \centering
    \resizebox{\linewidth}{!}{
    \begin{tabular}{l|ccc|ccc|c}
        \toprule 
        {\multirow{2}*{\textbf{\textsc{Model}}}} & \multicolumn{3}{c|}{\textbf{\textsc{TabFact}}} & \multicolumn{3}{c|}{\textbf{\textsc{WikiTableQuestion}}}& \multicolumn{1}{c}{\textbf{\textsc{FeTaQA}}} \\
        & All & Simple & Complex & All & Simple & Complex & BLEU  \\
        \midrule
            \textbf{\textsc{Dater}} & \textbf{85.6} & \textbf{91.2} & \textbf{80.0} & \textbf{65.9} & \textbf{68.2}& \textbf{63.5}& \textbf{30.92} \\
            \quad w/o \textit{Evidence Decomposition} & 81.8 ($\downarrow$ 3.8) & 86.9 ($\downarrow$ 4.3) & 76.8 ($\downarrow$ 3.2) & 63.9 ($\downarrow$ 2.0) & 65.5 ($\downarrow$ 2.7)& 62.2 ($\downarrow$ 1.3) & 28.46 ($\downarrow$ 2.46)  \\
            \quad w/o \textit{Question Decomposition} & 81.9 ($\downarrow$ 3.7) & 90.0 ($\downarrow$ 1.2) & 74.1 ($\downarrow$ 5.9) & 61.4 ($\downarrow$ 4.5)& 63.6 ($\downarrow$ 4.6) & 59.1 ($\downarrow$ 4.4)& 30.73 ($\downarrow$ 0.19) \\
        \bottomrule
    \end{tabular}
    }
    \caption{Ablation results on the test sets of the three datasets.}
    \label{tab:main-ab}
\end{table*}

\subsection{Ablation Study}
To analyze the impact of two kinds of decomposition in Dater, we conduct an ablation study by discarding the evidence decomposition module (denoted as w/o \textit{Evidence Decomposition}) and the question decomposition module (denoted as w/o \textit{Question Decomposition}) on TabFact, WikiTableQuestion and FetaQA. The questions in the TabFact and WikiTableQuestion datasets can be further divided into two levels based on the difficulty of the questions:  \textit{Simple} and \textit{Complex}, which can be used to better evaluate the model performance on different questions. 
Specifically, TabFact is divided based on officially provided question difficulty labels, while WikiTableQuestion is categorized by the length of the question.  
While FetaQA questions are more similar in length and difficult to distinguish, so we do not divide them.
The ablation test results are reported in Table~\ref{tab:main-ab}. 
It is no surprise that combining both evidence and question composition achieves the best performance in all the experimental settings.
For FabFact, both evidence decomposition and question decomposition have large impacts on the performance of Dater. The accuracy of Dater without evidence/question decomposition decreases by 3.8\%/3.7\% on the \textit{All} test set of FabFact. For WikiTableQuestion, question decomposition has a much larger impact than evidence decomposition on the performance of Dater. This is because WikiTableQuestion contains more complex questions that involve comparison, aggregation and arithmetic operations. This verifies the effectiveness of our question decomposition method in dealing with complex questions. 
For FetaQA, the questions are simpler and involve less numerical reasoning, thus the performance bottleneck is mainly in table content understanding. The performance of Dater drops significantly when discarding evidence decomposition. 
As FetaQA primarily focuses on the data-to-text generation, it is within our expectation that there is no significant gain from question decomposition. 

\begin{table*}
\resizebox{\linewidth}{!}{
\begin{tabular}{lll}
\toprule
\rowcolor[RGB]{240,248,255} \textsc{\textbf{Semantic Understanding}} &&\\
\midrule
Question & on august 25 , remlinger (6 - 6) took the win against the rockies.                                \\
Rule-based method & date, loss    & \sqlwrong{}               \\
Dater & date, loss, score, opponent & \sqlcorrect{}      \\
Ground Truth: & date, loss, score, opponent  &      \\
\midrule
\rowcolor[RGB]{240,248,255} \textsc{\textbf{Commonsense Knowledge}} &&\\
\midrule
Question &  the top scorer (matias suarez) has 5 fewer goals than  bart goor , who has 11 goals.&\\
Rule-based method &  player, league &\sqlwrong{}          \\
Dater &  player, total   & \sqlcorrect{}           \\
Ground Truth: &  player, total    &      \\
\midrule
Question &  denver did not lose more than one game in a row during november. & \\
Rule-based method &  date    & \sqlwrong{}            \\
Dater &  team, date, score & \sqlcorrect{}  \\
Ground Truth & team, date, score  &  \\
\bottomrule
\end{tabular}
}
\caption{Case study on evidence decomposition results predicted by a rule-based method and Dater. We categorize the questions according to the knowledge required for decomposition, including semantic understanding and commonsense knowledge.}
  \label{tab:col_select}
\end{table*}

\begin{table*}[h]
  \resizebox{\textwidth}{!}{
  \begin{tabular}{ll}
    \toprule
    \rowcolor[RGB]{240,248,255} \textsc{\textbf{WikiTableQuestion}}  & \\
    \midrule
    Question 1    & what is the number of listings from barrington, farmington, and rochester combined?  \\
    Codex CoT   &   there are totally 1+1+1=3 listings from barrington, farmington, and rochester combined.\\
    Codex Prediction &  ``3'' \sqlwrong{}  \\
    Dater Sub-question  & \makecell[l]{the number of listings from barrington is \{...\}. the number of listings from farmington is \{...\}. \\ the number of listings from rochester is \{...\}.} \\
    Dater Parsing & \makecell[l]{SELECT COUNT(*) FROM w WHERE city or town = 'barrington' \\ SELECT COUNT(*) FROM w WHERE city or town = 'farmington' \\ SELECT COUNT(*) FROM w WHERE city or town = 'rochester'}\\
    Dater CoT  & \makecell[l]{the number of listings from barrington is \{1\}. the number of listings from farmington is \{1\}. \\ the number of listings from rochester is \{3\}.} \\
    Dater Prediction   &   ``5'' \sqlcorrect{} \\
    Gold Answer   &   ``5''  \\
    \midrule
    Question 2   &  does the team have more 5th or 7th place finishes during the regular season? \\
    Codex CoT   &  the team have 1 5th place finishes and 2 7th place finish during the regular season. \\
    Codex Prediction   &  ``7th'' \sqlwrong{} \\
    Dater Sub-question  &  the team have \{...\} 5th place finishes during the regular season. the team have \{...\} 7th place finishes during the regular season. \\
    Dater Parsing & \makecell[l]{ SELECT COUNT(*) FROM w WHERE 
    reg.season LIKE '\%5th\%'  \\ SELECT COUNT(*) FROM w WHERE reg.season LIKE '\%7th\%' }\\
    
    Dater CoT  &  the team have \{3\} 5th place finishes during the regular season. the team have \{2\} 7th place finishes during the regular season. \\
    Dater Prediction  &   ``5th'' \sqlcorrect{} \\
    Gold  Answer  &   ``5th''\\
    \midrule
    \rowcolor[RGB]{240,248,255} \textsc{\textbf{TabFact}}  &  \\
    \midrule
    Question 1   & pádraig harrington and graeme mcdowell are both fron northern ireland.   \\
    Codex CoT  &  pádraig harrington is fron northern ireland.graeme mcdowell is fron northern ireland. \\
    Codex Prediction &  True \sqlwrong{} \\
    Dater Sub-question  &  pádraig harrington is \{...\} fron northern ireland. graeme mcdowell is \{...\} fron northern ireland. \\
    Dater Parsing & \makecell[l]{
    SELECT COUNT(*) FROM w WHERE player = 'pádraig harrington' AND country = 'northern ireland'\\
    SELECT COUNT(*) FROM w WHERE player = 'graeme mcdowell' AND country = 'northern ireland'}\\
    Dater CoT  &  pádraig harrington is \{0\} fron northern ireland. graeme mcdowell is \{1\} fron northern ireland. \\
    Dater Prediction &  False \sqlcorrect{} \\

    Gold Answer   &  False \\
    \midrule
    Question 2 & atlético ciudad played 28 matches with an average of less than 0.61 
    \\

    Codex CoT  & atlético ciudad played 28 matches. the average of atlético ciudad is 0.61.  \\
    Codex Predict   &   True \sqlwrong{}    \\
    Dater Sub-question  &   atlético ciudad played \{...\} matches. atlético ciudad played \{...\} matches with an average of less than 0.61.\\
    Dater Parsing &  \makecell[l]{SELECT SUM(matches) FROM w WHERE team = 'atlético ciudad'\\SELECT COUNT(*) FROM w WHERE team = 'atlético ciudad' AND average < 0.61 }     \\
    Dater CoT  &   atlético ciudad played \{28\} matches. atlético ciudad played \{0\} matches with an average of less than 0.61.\\
    Dater Prediction & False  \sqlcorrect{} \\
    Gold Answer   & False \\
    \bottomrule
  \end{tabular}}
  \caption{
  Case study on question decomposition results predicted by Codex and Dater. We also show the intermediate parsing results, which have the potential to provide the advantage of interpretability.
  }
  \label{tab:cs_cot}
\end{table*}

\subsection{Case Study}
\paragraph{Case Study for Evidence Decomposition}
To better understand how evidence decomposition helps our model capture the relevant sub-evidence, we demonstrate 
three exemplary cases about sports selected from TabFact. Due to space limitations, we solely provide the table column selection results. From Table~\ref{tab:col_select}, we can observe that the rule-based method fails to precisely align the question and evidence since it cannot capture the deep semantic information of both the question and evidence. Taking the first case as an example, the rule-based method cannot link the phrase ``\textit{against the rockies}'' and the column name ``\textit{opponent}''.  In addition, the rule-based method often suffers from a lack of commonsense knowledge and cannot comprehend complex questions effectively. For example, as shown by the third case, the rule-based method fails to recognize ``\textit{denver}'' as a team name and does not discover that the ``\textit{score}'' column can be used to determine the results of the games. 
On the contrary, our Dater method can address the limitations of previous methods in evidence decomposition and accurately select the relevant columns from a table by eliciting rich semantic and commonsense knowledge from a powerful LLM. 

\begin{wrapfigure}{l}{0.5\textwidth}
\begin{tikzpicture}
\begin{axis} [
ybar,
height=1.2in, 
bar width=0.4cm,
width=0.95\linewidth,
scale only axis,
ymin = 0, 
yticklabels=\empty,
axis x line*=bottom,
hide y axis,
xticklabel style = {font=\small,yshift=0.5ex},
symbolic x coords={
TabFact,
WikiTableQuestion,
FetaQA,
},
legend style={
    at={(0,1.0)},
    anchor=north west,
    legend columns=-1
},
enlarge x limits=0.2,
xtick=data,
yticklabels=\empty,
nodes near coords,
nodes near coords align={vertical},
every node near coord/.append style={font=\small},
legend style={at={(1,1)},anchor=north east}
]
\addplot coordinates {
(TabFact, 94) (WikiTableQuestion, 164) (FetaQA, 90)
};
\addplot coordinates {
(TabFact, 30) (WikiTableQuestion, 56) (FetaQA, 32)
};
\legend{Before, After}
\end{axis}
\end{tikzpicture}
\caption{Comparison of the average number of table cells before and after evidence decomposition on three dataset.}
\label{fig:table_cell}
\end{wrapfigure}

\paragraph{Case Study for Question Decomposition}
We use four exemplar cases selected from WikiTableQuestion and TabFact datasets to verify the effectiveness of our question decomposition qualitatively.
Table~\ref{tab:cs_cot} shows the ``chain of thought'' generated by Codex and our Dater method, where the first two cases are from WikiTableQuestion and the last two cases are from TabFact. From the cases, we can observe that Dater can decompose the complex question into simpler sub-questions that can be easily solved by leveraging a ``parsing-execution-filling'' strategy. In particular, Dater can generate high-quality SQL queries to retrieve correct information from the evidence. The retrieved information is essential to predict the final answer.  In contrast, without question composition, Codex fails to understand and comprehend complex questions, therefore it just gets information from the given question. For example, Codex tends to generate a ``\textit{chain of thought}'' (a series of intermediate reasoning steps) that is inconsistent with the evidence.

\subsection{Analysis of Sub-evidence}
As revealed by \citep{llmtab}, the table-based reasoning models are unable to generalize to ``huge'' tables with 30+ rows, which is the major error source. To address this problem, evidence decomposition is proposed to obtain a sub-evidence by filtering the irrelevant information from the ``huge'' evidence. To illustrate the effectiveness of evidence decomposition in narrowing the scope of irrelevant evidence, in Figure \ref{fig:table_cell}, we report the average number of table cells before and after evidence decomposition on the three datasets. From the results, we can observe that the sub-evidence is about 3x smaller than the original evidence, while achieving better performance. In particular, for WikiTableQuestion, we can reduce the average number of table cells from 164 to 56, significantly relieving the burden of LLMs in dealing with ``huge'' tables. Similar trends can also be observed on TabFact and FetaQA datasets.

\section{Conclusion}
In this paper, we explored in-context learning in LLMs to decompose structured evidence and unstructured NL questions for effective table-based reasoning. First, we utilized a powerful LLM to decompose the evidence involved in the current question into a relevant sub-evidence. The sub-evidence extraction could identify the relevant sub-evidence and exclude the remaining irrelevant evidence by predicting the indexes of the rows and columns with the help of an LLM and a few prompts.
Second, we introduced a ``\textit{parsing-execution-filling}'' strategy, which explored the programming language SQL to decompose the complex question into logical and numerical computation. 
Finally, we leveraged the decomposed sub-evidence and sub-questions to obtain the final answer effectively with the help of a few in-context prompting examples.   
Experimental results on three benchmark datasets demonstrated that our Dater achieved significantly better results than previous competitive baselines (fine-tuning-based methods and LLM-based in-context learning methods) for table-based reasoning. Particularly, Dater outperforms human performance for the first time on the TabFact dataset. In addition to impressive overall performance, Dater also has the advantage of interpretability, where the returned results are to some extent tractable with the generated sub-evidence and sub-questions. 

Our evidence decomposition method extracts indexes of the rows and columns as a whole via a powerful LLM without considering the ``chain-of-thought'' characteristic of the given question. In the future, we plan to learn the fine-grained alignment between tables and questions by performing step-by-step reasoning over tables.

\bibliographystyle{acl_natbib}
\bibliography{main}


\end{document}